\documentclass[letterpaper, 10 pt, conference]{ieeeconf}  % Comment this line out if you need a4paper

\IEEEoverridecommandlockouts                              % This command is only needed if 
\usepackage{caption}
\usepackage{multirow}
\usepackage{subfigure}
\usepackage{siunitx}
\usepackage{ulem} 
\usepackage{amsmath,amsfonts,amssymb,mathrsfs}  % assumes amsmath package installed
\usepackage{float} % 在导言区添加
\usepackage{bm}
\usepackage{ulem}
\usepackage{mathtools}
\usepackage{booktabs}
\usepackage{graphicx}
\usepackage{subcaption}
\makeatletter  
\newif\if@restonecol  
\makeatother

\usepackage[linesnumbered,ruled,vlined]{algorithm2e}%[ruled,vlined]
\usepackage{algpseudocode}  
\usepackage[colorlinks,urlcolor=blue,linkcolor=blue,citecolor=blue]{hyperref}
\newtheorem{theorem}{Theorem}[section]

\newtheorem{lemma}[theorem]{Lemma}

\newcommand\norm[1]{\lVert#1\rVert}

\title{\LARGE \bf
Online Time-Informed Kinodynamic Motion Planning of Nonlinear Systems}

\author{Fei Meng, Jianbang Liu, Haojie Shi, Han Ma, Hongliang Ren,~\IEEEmembership{Senior Member,~IEEE}, and \\Max Q.-H. Meng,~\IEEEmembership{Fellow,~IEEE} % <-this % stops a space
\thanks{This work was supported in part by Hong Kong RGC CRF grant C4063-18G and National Natural Science Foundation of China grant \#62103181. (\textit{Corresponding authors: Hongliang Ren and Max Q.-H. Meng})}% <-this % stops a space
\thanks{Fei Meng, Jianbang Liu, Haojie Shi, Han Ma, and Hongliang Ren are with the Department of Electronic Engineering, The Chinese University of Hong Kong, Hong Kong, {\tt\small \{feimeng, henryliu, h.shi, hanma\}@link.cuhk.edu.hk; hlren@ieee.org}}%
\thanks{Max Q.-H. Meng is with Shenzhen Key Laboratory of Robotics Perception and Intelligence and the Department of Electronic and Electrical Engineering at Southern University of Science and Technology in Shenzhen, China, and also a Professor Emeritus in the Department of Electronic Engineering at The Chinese University of Hong Kong, Hong Kong, {\tt\small max.meng@ieee.org}}
}

\begin{document}

\maketitle
\pagestyle{empty}  % no page number for the second and the later pages
\thispagestyle{empty} % no page number for the first page

%%%%%%%%%%%%%%%%%%%%%%%%%%%%%%%%%%%%%%%%%%%%%%%%%%%%%%%%%%%%%%%%%%%%%%%%%%%%%%%%
\begin{abstract}
Sampling-based kinodynamic motion planners (SKMPs) are powerful in finding collision-free trajectories for high-dimensional systems under differential constraints.
Time-informed set (TIS) can provide the heuristic search domain to accelerate their convergence to the time-optimal solution.
% The ellipsoidal technique has been used to approximate TIS under an HJB reachability formulation.
However, existing TIS approximation methods suffer from the curse of dimensionality, computational burden, and limited system applicable scope, e.g., linear and polynomial nonlinear systems.
To overcome these problems, we propose a method by leveraging deep learning technology, Koopman operator theory, and random set theory. 
Specifically, we propose a Deep Invertible Koopman operator with control U model named DIKU to predict states forward and backward over a long horizon by modifying the auxiliary network with an invertible neural network.
A sampling-based approach, ASKU, performing reachability analysis for the DIKU is developed to approximate the TIS of nonlinear control systems online.
Furthermore, we design an online time-informed SKMP using a direct sampling technique to draw uniform random samples in the TIS.
Simulation experiment results demonstrate that our method outperforms other existing works, approximating TIS in near real-time and achieving superior planning performance in several time-optimal kinodynamic motion planning problems.

% a steering function is not always easy and linearization is valid only locally, which causes. To address this issue, we derive a globally linearized dynamic system for a nonlinear system by ultilizing Koopmam operators.    
% task constraint is treated as convex nonlinear terminal constraint which finds multiple converging solutions via solving Euler-Lagrange equations. RRT* algorithm operates in the task space? 
% compute the optimal trajectory $\pi^*[\bm{\bm{x}_0},x_1]$ between any two states $\bm{x}_0 \in \mathscr Q_{safe}$ and $x_1 \in \mathscr Q_{safe}$ considering dynamics. Specically, calculate optimal robot state trjectory between any states at first. Then, determine the feasible trajectory $\bm{x}(t)$ with RRT*.
% Note that although we know the nonlinear differential equations for each system, we only used simulation data to perform Koopman linearization, treating each system as a black box. Since few approaches can work directly with blackbox systems, this allows us to compare against other reachability methods that require knowing the system’s dynamics.
% keyword: sampling based, kinodynamic motion planning problem with task space constraints for 7 DoFs redundant manipulator; finite time horizon optimal control
\end{abstract}

%%%%%%%%%%%%%%%%%%%%%%%%%%%%%%%%%%%%%%%%%%%%%%%%%%%%%%%%%%%%%%%%%%%%%%%%%%%%%%%%
\section{INTRODUCTION}
%\cite{lew2021sampling} requires a nominal trajectory and model information.
%Control Lyapunov \cite{zinage2023neural} (and Barrier) functions are also used to design safety controller [28, 29]; however, finding feasible Lyapunov functions can be challenging, especially in the presence of control or disturbance bounds.
%如果能求出可达集中与终点有关的密度，比以前IRRT和tang的更好
%Hamilton-Jacobi (HJ) reachability analysis can compute reachable sets exactly [21, 22]. Unfortunately, handling arbitrary systems (e.g., neural networks) with such approaches is challenging, as they require solving a partial differential equation involving a max/minimization over controls and disturbances. For this reason, these methods typically discretize the state space and thus suffer from the curse of dimensionality [21]. Further, these methods leverage the principle of dynamic programming to compute solutions, and thus cannot handle parameter uncertainty which introduces time correlations along the trajectory (see Section 2). computing the reachable states is proved to be undecidable in general (e.g., polynomial dynamical systems with degrees larger than 2) [24] and is also empirically time-consuming, limiting applications to simple dynamics (e.g., linear systems) or low-dimension systems.
Sampling-based motion planning methods are known for rapidly finding collision-free ``geometric" paths from a start to a goal point in high-dimensional C-space.
% Compared to probabilistic roadmap \cite{kavraki1996probabilistic} that builds a roadmap and search, 
The popular RRT \cite{lavalle2001randomized} grows a tree structure by drawing random samples until reaching the goal, guaranteeing probabilistic completeness.
% , i.e., a solution will be found eventually if it exists, as the number of iterations goes to infinity.
Its optimal version, RRT$^*$ \cite{karaman2011sampling}, provides the property of asymptotic optimality, whereas it costs too much time due to a uniform sampling strategy in the whole C-space \cite{meng2022nr,meng2023learning}.
To speed up the convergence to the optimal solution, some heuristic sampling approaches have been proposed; for example, ``$L_2$ Informed Set" containing all the potential shorter solutions is constructed to restrict the subsequent searches after an initial path is found \cite{gammell2014informed}. 
\begin{figure}[htbp]
\centering
\includegraphics[width=0.48\textwidth]{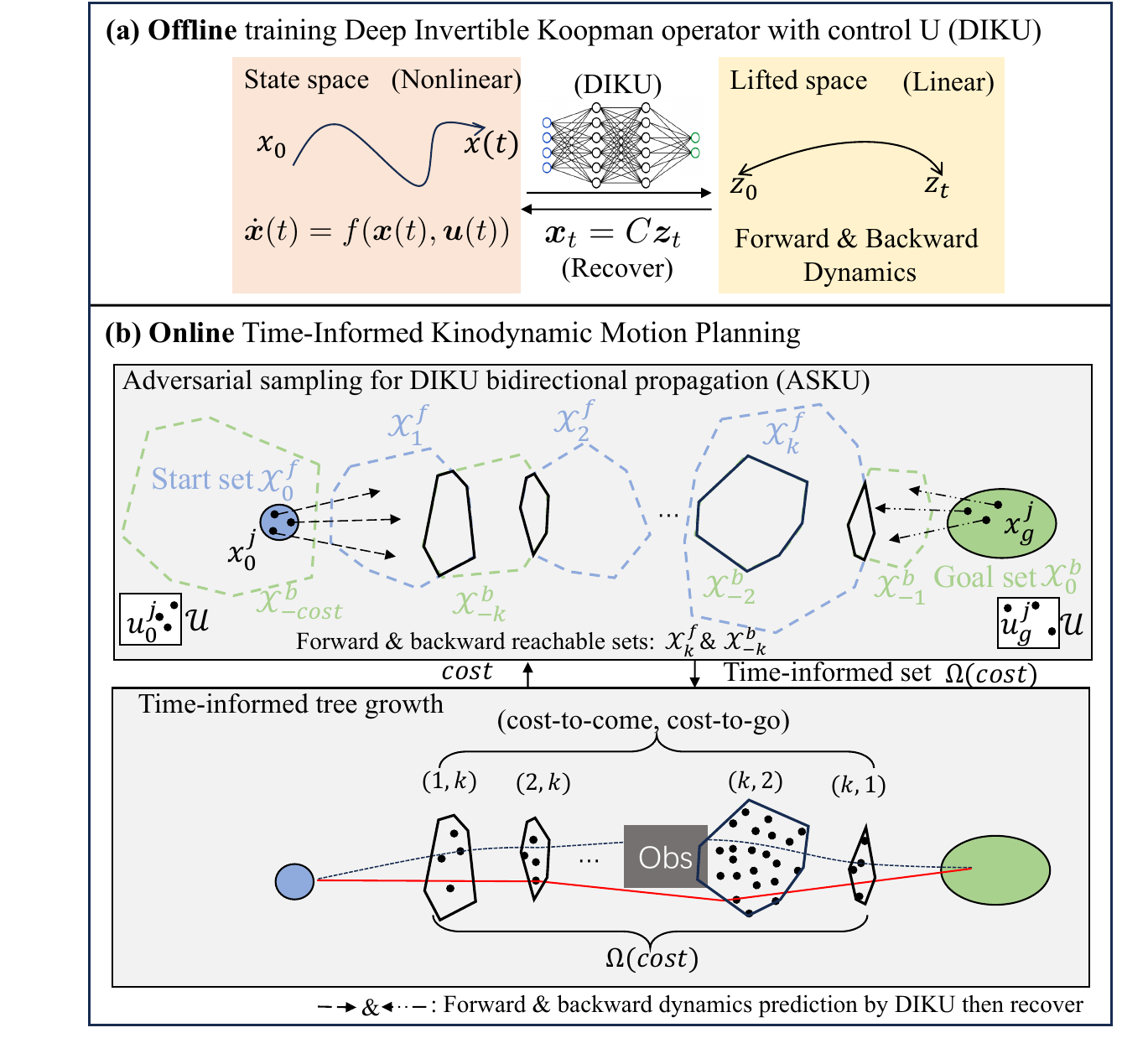}
\caption{An illustration of online time-informed kinodynamic motion planning of nonlinear control systems. 
(a) A Deep Invertible Koopman operator with control U model, DIKU, is trained offline for the nonlinear systems to obtain equivalent linear systems that enable forward and backward dynamics prediction in the lifted space.
(b) Our algorithm randomly samples states in the start and goal sets. It then uses the ASKU to bidirectionally propagate the learned linear dynamics and then recover, generating the forward reachable set $\mathcal{X}_k^f$ (independent blue dashed convex hull) and backward reachable tube $\mathcal{X}^b_{-cost:0}$ (serials of green dashed convex hulls) in near real-time.
Their intersections constitute the time-informed set (TIS) $\Omega(cost)$ (black convex polygons). 
Time-informed tree growth is achieved for the off-the-shelf SKMPs by directly sampling in the TIS.
The TIS updates according to the cost of the search tree returned after sufficient iterations to help refine the solution (red line).
}
\label{fig:about}
\end{figure}
When kinodynamic constraints, i.e., kinematic and differential constraints, are considered, the problem becomes PSPACE-hard and imposes a heavy computational burden.
Optimal trajectories under kinodynamic constraints, instead of line segments, connecting any two intermediate tree nodes are required to form the edges of the tree.  
% , which is NP-hard.
% and fails if the local target is not reached.
% Local trajectory optimization (e.g., linear quadratic regulator (LQR) RRT$^*$ \cite{perez2012lqr} and kinodynamic RRT$^*$ \cite{6631299}) and the methods of random shooting in control space (e.g., Stable Sparse RRT (SST) \cite{li2016asymptotically} and \cite{lavalle2001randomized}) are widely used.
Local trajectory optimization solvers \cite{perez2012lqr,6631299} and random shooting methods \cite{li2016asymptotically,lavalle2001randomized} are widely implemented to solve this two-point boundary value problem (TPBVP).
% For instance, \cite{perez2012lqr} linearizes the system dynamics about newly sampled node and use the LQR cost function to select the ``nearest" node on the tree and solve the LQR problem for steering.
% For instance, \cite{perez2012lqr} linearizes the dynamics and uses the linear quadratic regulator (LQR) cost function to select the ``nearest" node on the tree and solve the LQR problem for local steering.
% However, the solver's feasibility and the high probability of failure to satisfy the boundary constraints are bottlenecks.
Stable Sparse RRT (SST) \cite{li2016asymptotically} performs forward dynamic simulation with randomly sampled control signals for steering, eliminating the need for TPBVP solvers. 
Nevertheless, it is computation-demanding since its unbiased exploration of the whole state space.
% However, the distance function hinders efficient exploration due to the limitations of local linearization.
% The time-consuming computation and high probability of failure to satisfy the boundary constraints are also bottlenecks for local extension.
% Utilizing model predictive control (MPC) as a steering function is computationally efficient and able to provide the means to handle state or control constraints and return a near-optimal local trajectory even though boundary conditions are not satisfied \cite{li2021mpc}.
% However, the slow exploration and convergence to the optimality happens if we cannot design a proper domain-specific distance metric to determine the initial state of MPC at the absence of an iterative neural network sampler, which is a common case.
Hierarchical rejection sampling (HRS) \cite{kunz2016hierarchical} and Markov Chain Monte-Carlo (MCMC) \cite{yi2018generalizing} algorithms were proposed to 
generate samples within the informed set of non-Euclidean state spaces.
However, the HRS method takes a longer time as the interested subspace shrinks or in high-dimensional space.
Although the Hit-and-Run (HNR) MCMC method \cite{yi2018generalizing} is efficient, it requires an initial solution and applies to only certain systems for minimum-time problems.

Reachability-guided methods have been proposed to reduce computation cost \cite{shkolnik2009reachability}.
\textcolor{black}{At each extension, the polytopic forward reachable set (FRS) of each state is approximated to avoid selecting the nearest but unreachable state \cite{wu2020r3t}.
In addition, the TIS that includes all feasible kinodynamic trajectories is provided to restrict the sampling domain \cite{joshi2021tie}. In principle, a set operation of the FRS of the start and the backward reachable tube (BRT) of the goal state is conducted. Tang \textit{et al.} \cite{tang2022reachability}, which is most relevant to ours, relaxed the Hamilton-Jacobi-Bellman (HJB) equation to compute a modified TIS, broadening the system scope from linear \cite{joshi2021tie} to polynomial nonlinear (PNL) systems and conducting a heuristic search from the beginning. However, it suffers from the curse of dimensionality, a heavy computational burden, and limited system-specific applicable scope.
The sampling-based reachability analysis method that takes convex hulls for propagated states with random control inputs from the start has great potential to solve these problems by leveraging random set theory \cite{molchanov2005theory}.
However, only FRSs can be approximated since there is no generic differential equation for the backward propagation of nonlinear systems. Considering Deep Koopman Operator (DKO) can predict the forward evolution of nonlinear dynamics by mapping the original system into an equivalent linear one in an observation space via deep neural network (DNN) \cite{shi2022deep} and linear systems are invertible, we integrate it into the sampling-based method to pave the path of constructing BRT. Furthermore, an adversarial sampling manner can guarantee the over-approximation reachability analysis, benefiting informed sampling. Although reachable tubes for high-dimensional nonlinear systems can be approximated via DNN end-to-end, it takes long computation times from hours to days \cite{bansal2021deepreach}. By contrast, our method only needs a short time to learn the locomotion characteristics of the nonlinear systems to realize reachability analysis indirectly.
}

% On the other hand, Thapliyal \textit{et al.} \cite{thapliyal2023approximating} learned the system's DNN first and then deliberately utilized a data-driven method to derive the Koopman operator (KO) with control. 
% The polytopic approximations of FRSs for quadrotors are finally derived in real time by applying optimal control theory.

% \cite{bak2021reachability} computed the zonotope reachable sets after Koopman operator linearization.
% However, it is only applicable to non-control dynamical systems and is computationally expensive.
% The off-the-shelf local steering functions in these reachability guided SKMPs also inherit the drawbacks mentioned above.

In this letter, we leverage the deep learning, KO and random set theories to contribute a generic and efficient time-informed SKMP.
The paradigm of the planning procedure is shown in Fig. \ref{fig:about}.
The main contributions are as follows\footnote{Source code: \href{https://github.com/feimeng93/OnlineTimeInformedKinoMP}{https://github.com/feimeng93/OnlineTimeInformedKinoMP}}. 
\begin{itemize}
    \item A deep invertible Koopman operator with control U, named DIKU, that can bidirectionally predict the long-term evolution of nonlinear control systems is proposed by integrating the invertible neural networks (INN) \cite{dinh2016density} into our previous DKO model \cite{shi2022deep}.
    % We additionally train an auxiliary network to generate the previous states given batch current states and control signals, leading to a Prediction \& Recall Deep Koopman Operator (PRDKO) that can not only predict but recall states under kinodynamic constraints.
    \item TIS is approximated online for the nonlinear control systems \textcolor{black}{that can be learned}. We use a sampling-based method \cite{lew2021sampling} to quickly perform the reachability analysis based on the bidirectional propagation of DIKU, overcoming the curse of dimensionality, broadening the system's applicable scope and boosting the computation efficiency for the TIS significantly.
    \item We develop an online time-informed kinodynamic motion planning algorithm where direct sampling in the TIS is realized. We evaluate our method on six different types of dynamic systems and achieve better planning efficiency over the existing works.

% A collision-aware deep koopman operator model predictive controller (MPC) for local extension is proposed. 
% It accounts for not only state but control constraint satisfaction and optimality, compared to \cite{tang2022reachability} where only control constraints are considered.
% More accurate distance measure and non-collapsed local optimal trajectories are realized in the koopman lifted linear space without an exact known model, compared to \cite{li2021mpc}.
\end{itemize}

\section{Problem Formulation and Preliminaries}
\subsection{Time-optimal Kinodynamic Motion Planning}
Let compact sets $\mathcal{X}\subset\mathbb{R}^{n}\,(n\geq1)$ and $\mathcal{U}\subset\mathbb{R}^{m}\,(m\geq1)$ be state and admissible control spaces, respectively.
Denote $\mathcal{X}_{\text{obs}}\subseteq\mathcal{X}$ as the obstacle space and the closure of the set difference $\mathcal{X}_{\text{free}}=\text{cl}(\mathcal{X}\backslash\mathcal{X}_{\text{obs}})$ as the free space.
Let $\bm{x}_0\in\mathcal{X}_{\text{free}}$ be the start state and $\mathcal{X}^g\subset\mathcal{X}_{\text{free}}$ be the goal set.
Let us define a dynamical system as follows.
\begin{equation}
\dot{\bm{x}}(t)=f(\bm{x}(t),\bm{u}(t)),
\label{eq:dynamic model}
\end{equation}
where $t$ is the sampling time, $\bm{x}(t)\in\mathcal{X}$ is the state, $\bm{u}(t)\in\mathcal{U}$ is the control signals, and the  flow field $f(\cdot):\mathcal{X}\times\mathcal{U}\rightarrow\mathcal{X}$ is continuously differentiable.

% \begin{assumption}[Lipschitz Continuity] 
% $\left.1\right)$ The state space $\mathcal{X}$ and admissible control space $\mathcal{U}$ are compact. 

% $\left.2\right)$ $f(\cdot)$ is uniformly continuous, bounded, and Lipschitz in $\bm{x}\in\mathcal{X}$ for every $\bm{u}\in\mathcal{U}$.

% $\left.3\right)$ The target set $\mathcal{X}^g$ is closed and can be represented as the subzero level set of a bounded and Lipschtiz function $g(\cdot):\mathbb{R}^{n}\rightarrow\mathbb{R}$, namely  $\mathcal{X}^g=\left\{\bm{x}\in\mathbb{R}^{n}\,|\,g(\bm{x})\leq 0\right\}$.
% \end{assumption}

Given an initial state  $\bm{x}_0\in\mathcal{X}_{\text{free}}$ and a target set $\mathcal{X}^g\subset\mathcal{X}_{\text{free}}$, we aim to determine the optimal control signals $\bm{u}$ and minimum time-to-reach $t_f$ such that a time-optimal trajectory $[\bm{x}(0),\dots,\bm{x}(t_f)]$ connecting the start and a goal state $\bm{x}(t_f)\in\mathcal{X}^g$ is found for \eqref{eq:dynamic model}.
Mathematically, we have the time-optimal kinodynamic motion planning problem:
\begin{equation}
\begin{aligned}
	&\hspace{25mm}\mathop{\min}_{\bm{u}}t_f\\
		\text{s.t.}\,&\dot{\bm{x}}(t)=f(\bm{x}(t),\bm{u}(t))\\
		&\bm{x}(0) = \bm{x}_{0}, \bm{x}(t_f)\in\mathcal{X}^g, t_f>0,\text{and}\; t_f\;\text{is free}\\
		&\bm{x}(t)\in\mathcal{X}_{\text{free}},\, \bm{u}(t)\in\mathcal{U},\forall t \in [0, t_f].
\end{aligned}
\label{eqoptimalcost}
\end{equation}

\subsection{Time-Informed Set}
The forward reachable set (FRS) is defined as the set of all possible states that can be reached at time $t$, starting from $\bm{x}_0$ at time $0$, using admissible controls,
\begin{equation}
\begin{aligned}
	&\mathcal{X}^f(0,t,\bm{x}_0)\triangleq \left\{\bm{x}_t\in \mathcal{X}\;|\;\forall \tau \in [0,t],\exists \bm{u}(\tau)\in \mathcal{U}\right. \label{eq:FRS}\\  
	&\text{s.t.}\,\left. \bm{x}(0)=\bm{x}_0,\bm{x}(t)=\bm{x}_t,\dot{\bm{x}}(\tau)=f(\bm{x}(\tau),\bm{u}(\tau))\right\}
\end{aligned}    
\end{equation}
The backward reachable set (BRS) for the target set $(t_f, \mathcal{X}^g)$ is the set of all states at time $t$ that will reach the goal at $t_f$,
\begin{equation}
\begin{aligned}
	& \mathcal{X}^b(t,t_f,\mathcal{X}^g)\triangleq \left\{\bm{x}_t\in \mathcal{X}\;|\;\forall \tau \in [t,t_f],\exists \bm{u}(\tau)\in \mathcal{U}\right.  \label{eq:BRS}\\  
	&\text{s.t.}\,\left. \bm{x}(t)=x_{t},\bm{x}(t_f)\in\mathcal{X}^g,\dot{\bm{x}}(\tau)=f(\bm{x}(\tau),\bm{u}(\tau))\right\}.
\end{aligned}
\end{equation}
% The backward reachable tube (BRT) $\mathcal{X}_b(t_f,t,\mathcal{X}^g)$ over the interval $[t,t_f]$ for the target set $(t_f, \mathcal{X}^g)$ is the set of all states starting at time t $\bm{x}(t)$ will reach the goal at any time within the time horizon 
% \begin{equation}
% \begin{aligned}
% &\mathcal{X}_b(t_f,t,\mathcal{X}^g)\triangleq \left\{\bm{x}_t\in\mathcal{X}|\exists\uline{\tau}\in[t,t_f],\forall\tau\in [t,\uline{\tau}],\exists \bm{u}(\tau)\in\mathcal{U}\right.  \label{eq:BRSII}\\
% &\text{s.t.}\,\left. \bm{x}(t)=x_{t},\bm{x}(\uline{\tau})\in\mathcal{X}^g,\dot{\bm{x}}(\tau)=f(\bm{x}(\tau),\bm{u}(\tau))\right\}??
% \end{aligned}
% \end{equation}

Let $\mathcal{F}(0,t,\bm{x}_0)$ and $\mathcal{B}(t,t_f,\mathcal{X}^g)$ be the over-approximations of FRS and BRS, respectively.
The BRT that contains the set of all states $\bm{x}_{t}$ that can reach $\mathcal{X}^g$ at  $\forall\tau\in[t,t_f]$ is $\mathcal{R}_b(t,t_f,\mathcal{X}^g)=\cup_{t\leq\tau\leq t_f}\mathcal{B}(t,\tau,\mathcal{X}^g)$.

The time-informed set (TIS) of time cost $t_f$ is \cite{joshi2021tie}:
\begin{equation}
\Omega(t_f)\triangleq\mathop\bigcup\limits_{t\in[0,t_f]}\mathcal{F}(0,t,\bm{x}_0)\cap\mathcal{R}_b(t,t_f,\mathcal{X}^g)\subset\mathcal{X}.
\label{eq:tis}
\end{equation}

\subsection{Koopman Invertible Autoencoder}
% The Deep Koopman Operator (DKO) can predict the forward evolution of nonlinear dynamical systems \eqref{eq:dynamic model} by mapping the original system into an equivalent linear one in the observation space as \eqref{eq:forward g} via DNN \cite{shi2022deep}.
A Koopman invertible autoencoder (KIA) is proposed by replacing the linear auxiliary network in the classic DKO with an INN architecture \cite{dinh2016density} to model both forward and backward dynamics \cite{tayal2023koopman}.
INN establishes a connection between the dual dynamics according to the bijective functions:
% , instead of training two independent networks as in \cite{azencot2020forecasting}
\begin{equation}
    \bm{g}_{t+1,1} = \bm{g}_{t,1} + f_2(\bm{g}_{t,2}),\;\bm{g}_{t+1,2} = \bm{g}_{t,2} +f_1(\bm{g}_{t+1,1});
    \label{eq:forward inn}
\end{equation}
\begin{equation}
    \bm{g}_{t-1,2}=\bm{g}_{t,2}-f_1(\bm{g}_{t,1}),\;\bm{g}_{t-1,1}=\bm{g}_{t,1}-f_2(\bm{g}_{t-1,2});
    \label{eq:backward inn}
\end{equation}
where $\bm{g}_t=[\bm{g}_{t,1};\,\bm{g}_{t,2}]$, $[;]$ is the vertical concatenation operation, and $f_1(\cdot)$ and $f_2(\cdot)$ are translation functions.
\textcolor{black}{The original lifted forward evaluation with KO $K$, $\bm{g}(\bm{x}_{t+1})=K\bm{g}(\bm{x}_t),\;\bm{g}(\bm{x})=[\psi_1(\bm{x}), \psi_2(\bm{x}),...,\psi_c(\bm{x})]^{T}\in\mathbb{R}^c\subset\mathcal{C}$},
can be approximated with \eqref{eq:forward inn}, and its inverse expression is obtained by \eqref{eq:backward inn} at no cost.
Although KIA can forecast and backcast temporal data, its evolution requires an NN decoder and does not include any control input \cite{tayal2023koopman}.
% Global feasible exploration tree $\mathcal T$ is conducted by the RRT* algorithm in the task space. Each node $\bm{n}^i$ of the tree consists of the desired task state $y(t_i)$ and the corresponding robot state $\bm{x}(t_i)=[q;\dot{q}]$ at a time instant $t_i$. Each edge $\mathcal E_{i.j}$ contains the desired task trajectory $\bm{y}_{i,j}(t)$ from $\bm{y}_i$ to $\bm{y}_j$, and the joint trajectory $\bm{q}(t)$ for $t \in [t_i,t_j]$. We use the kinodynamic optimal motion planner to optimize the rewire and parent modules of the RRT* algorithm. and the leaf method.

\section{Deep Koopman Reachability Analysis}
% This section proposes DIKU to model nonlinear control systems' long-term forward and backward dynamics. 
% We developed ASKU to approximate the systems' TIS online by leveraging KO and random set theories.

\subsection{Deep Invertible Koopman Operator with Control Enabling Long-Term Forward and Backward Propagation}

It has been proven that the DNN autoencoder can discover the function observables with linear evolution for controlled lifted linear system \cite{shi2022deep}. We parameterize the observable $\bm{z}(\bm{x}_t)$ as $\bm{z}_{\theta_z}(\bm{x}_t)$ by using a DNN with parameters $\theta_z$, and concatenate the state and embedded feature together as the observable as $\bm{z}_t=[\bm{x}_t; \bm{z}_{\theta_z}(\bm{x}_t)]\in\mathbb{R}^{n+d}$ to realize authentic recovery to nonlinear states.
% \begin{equation}
% z_t=
% \begin{bmatrix}
% \bm{x}_t \\ z_{\theta}(\bm{x}_t)
% \end{bmatrix}
% \in\mathbb{R}^{n+d},
% \label{eq:zx} 
% \end{equation}
% which leads to a simple recovery as $\bm{x}_k=C\bm{z}_k$, where $C=[I_n\;\;0]\in\mathcal{Z}\subset\mathbb{R}^{n\times(n+d)}$.
Furthermore, we have the following $K$-steps feedforward prediction using NNs: 
\begin{equation}
    \begin{aligned}  \bm{z}_{t+k+1}=K_x\bm{z}_{t+k}&+K_u\bm{u}_{t+k},\;k=0,\cdots,K-1 \\
        % \bm{z}_t&=
        % \begin{bmatrix}
        %     \bm{x}_t\\\bm{z}_{\theta_z}(\bm{x}_t)
        % \end{bmatrix}\\
        \bm{x}_{t+k}&=C\bm{z}_{t+k},
    \end{aligned}
    \label{eq:kstepz}
\end{equation}
where $K_x\in\mathbb{R}^{c_x\times c_x}$ and $K_u\in\mathbb{R}^{c_x\times c_u},\,c=c_x+c_u,$ are fully connected layers with parameters $\theta_x$ and $\theta_u$, respectively.
Note that nonlinear state constraints $c(\bm{x}_t)\leq0:\mathbb{R}^{n}\rightarrow\mathbb{R}$ can be considered by adding them to $\bm{z}_t$.
For instance, if there exists a constraint $\sin(x_1)\leq0$ , we can let $z_t^{(n+d+1)}=sin(x_1)$ and enforce it no greater than 0 \cite{korda2018linear}.
Observing the linearity in the lifted space, we could have backcasted via the inverse matrix of $K_x$.
However, it is computationally expensive, typically $O(n^3)$.

\begin{figure*}[htbp]
\centering
\includegraphics[width=0.9\textwidth]{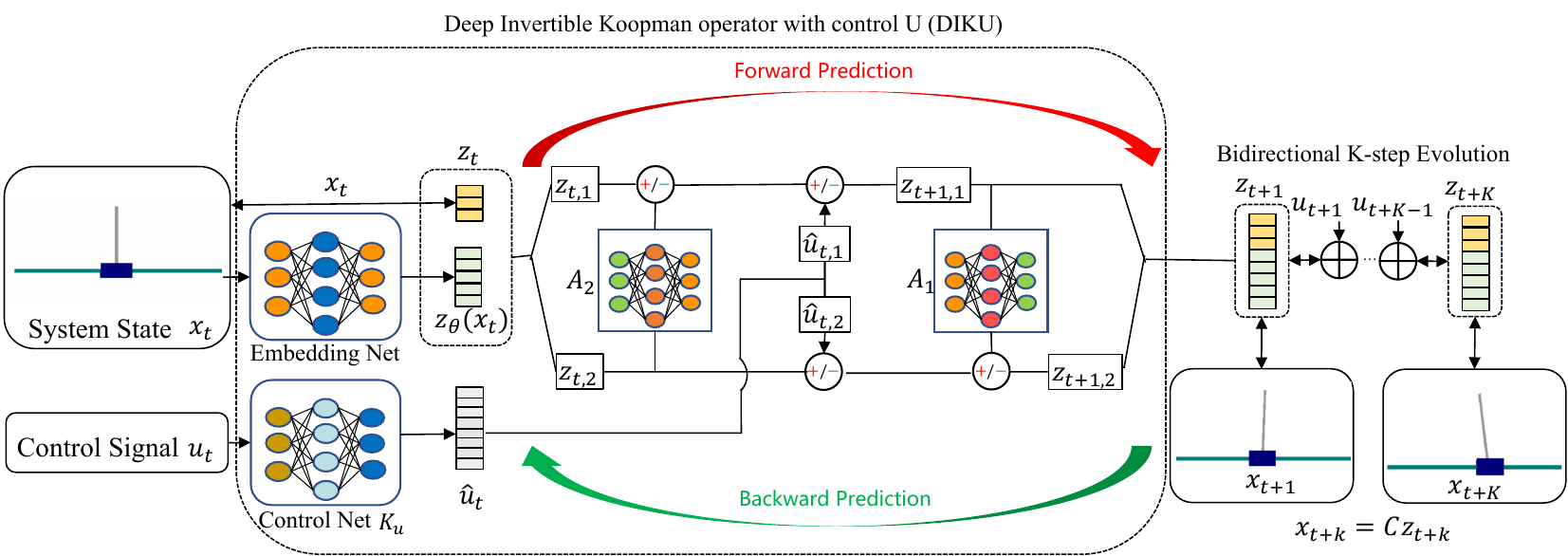}
\caption{An overview of Deep Invertible Koopman operator with control U (DIKU) neural network model for long-horizon forward and backward dynamics prediction
}
\label{fig:DKIU}
\end{figure*}

Herein, we propose the DIKU utilizing the invertible INNs to realize bidirectional dynamics, whose architecture is shown in Fig. \ref{fig:DKIU}. 
The latent features are divided into two parts as $\bm{z}_{t}=[\bm{z}_{{t},1};\,\bm{z}_{{t},2}]$ and so does the control field $K_uu_t=[B_1;B_2]u_t=\hat{\bm{u}}_t=[\hat{\bm{u}}_{t,1};\,\hat{\bm{u}}_{t,2}]$. 
Then, we parameterize the $f_1(\cdot)$ and $f_2(\cdot)$ by utilizing DNNs with parameters $\theta_1$ and $\theta_2$ followed by activation functions, and denote them as $A_1$ and $A_2$.
Consequently, \eqref{eq:kstepz} is rewritten as follows to improve long-term prediction ability:
\begin{equation}
\begin{aligned}
\bm{z}_{{t+k+1},1}&=\bm{z}_{{t+k},1}+A_2\bm{z}_{{t+k},2}+\hat{\bm{u}}_{{t+k},1},\\
\bm{z}_{{t+k+1},2}&=\bm{z}_{{t+k},2}+A_1\bm{z}_{{t+k+1},1}+\hat{\bm{u}}_{{t+k},2}.
\label{eq:FINN}
\end{aligned}
\end{equation}

The corresponding backward evolution in the lifted space is derived for free:
\begin{equation}
\begin{aligned}
\bm{z}_{{t-k-1},2}&=\bm{z}_{{t-k},2}-A_1\bm{z}_{{t-k},1}-\hat{\bm{u}}_{{t-k},2},\\
\bm{z}_{{t-k-1},1}&=\bm{z}_{{t-k},1}-A_2\bm{z}_{{t-k-1},2}-\hat{\bm{u}}_{{t-k},1}.
\label{eq:BINN}
\end{aligned}
\end{equation}

We collect the training trajectory dataset as $\mathcal{D}=\left[X_i\in\mathbb{R}^{N\times n},U_i\in\mathbb{R}^{N\times m},i=1,\cdots,K\right]$. 
% The real function observation is obtained by $Z_i=[X_i,z_{\theta}(X_i)]$. 
We define the following $K$-steps forward and backward dynamical loss as:
\begin{equation}    
\begin{aligned}
L(\theta_z,\,\theta_1,\theta_2,\,\theta_u)&=\frac{1-\gamma^K}{1-\gamma}\sum_{i=1}^K\gamma^{i-1}[MSE(X_{i},CZ_{i})\\
&+MSE(X_{K-i+1},CZ_{K-i+1})]
\end{aligned}
\label{eq:loss}
\end{equation}
where $MSE$ represents the mean square loss function and $\gamma$ denotes the weight decay constant. 
This $K$-steps loss is conductive to fine predictions in the long term.

% Three versions of $v(\cdot)$ have been implemented  (i.e., Deep KoopmanU with Control (DKUC), Deep Koopman Affine with Control (DKAC), and Deep Koopman Nonlinear Control (DKNC)).
% However, the DNN parameterization of DKNC needs the previous state and control signals as inputs, making the following equation infeasible
% \begin{equation}
% z_{t+k-1}=A^{-1}(z_{t+k}-Bv(x_{t+k-1},u_{t+k-1})).
% \label{eq:recall}
% \end{equation}
% and $v(x_k,u_k)=u_k$ for DKUC. $v(x_k,u_k)=v_{\phi_1}(x_k)u_k$ and $v(x_k,u_k)=v_{\phi_2}(x_k,u_k)$ represent DKAC and DKNC by using auxiliary control networks with parameters $\phi_1$ and $\phi_2$, respectively.
% To retrieve the previous function observables via \eqref{eq:recall}, we
% herein, only consider the DKU and DKAC with a simplified bilinear form (i.e., $v(x_k,u_k)=\{u_k,\,z_ku_k\}$).
% One layer linear network is used for $A^{-1}$ and the network of $B$ in prediction is reused to backward compute the function observables.

\subsection{Online Approximate Time-Informed Set via Random Set and Koopman Operator Theories}
\label{sec:kotis}
% The curse of dimensionality, computational cost, and system applicable scope limit the ellipsoidal technique \cite{tang2022reachability} to approximate the TIS of nonlinear systems.
% To overcome these problems, we propose the ASKU method by adversarial sampling for DIKU bidirectional propagation.

Random set theory \cite{molchanov2005theory} guarantees asymptotic convergence to the convex hull of the reachable sets, whose relevant theorem is restated as follows:
\begin{lemma}
    \cite{lew2021sampling}
Let $\left\{(\bm{x}_0^j,\bm{u}^j)\right\}_{j=1}^M$ be i.i.d. sampled parameters in $\mathcal{X}_0\times\mathcal{U}^{k-1}$, where $M$ denotes the number of states. Define $\bm{x}_t^j$ as \eqref{eq:dynamic model}. %$\bm{x}_{t+1}=f(\bm{x}_t,\bm{u}_t)$, where $f$ is continuously differentiable.
Assume that the sampling distribution of the parameters satisfies $\mathbb{P}(\bm{x}_t^j\in G_t)>0$ for any open set $G_t$ s.t. $\mathcal{X}_t\cap G_t\neq\emptyset$.
Then, as $M\rightarrow\infty$, $\mathcal{X}_t$ converges to the convex hull of the RS $\text{Co}(\mathcal{X}_t)$ almost surely.
\label{lemma1}
\end{lemma} 
% \begin{remark}
%     Since the theorem 2 in \cite{lew2021sampling} holds for any continuously differentiable dynamics, including neural network approximators and common robotic systems, we restate it in Lemma \ref{lemma1} except for uncertain parameters and disturbance that were considered.
% \end{remark}

A random set is defined as a map from a probability space to a family of sets.
Consider sampling i.i.d. initial tuples $(\bm{x}_0^j,\bm{u}_0^j)\in\mathcal{X}^f_0\times\mathcal{U}$ and goal tuples $({\bm{x}}_g^j,{\bm{u}}_g^j)\in{\mathcal{X}}^b_0\times\mathcal{U}$ according to arbitrary probability distribution, where $j=1,\cdots,M$.
Then, we encode the sampled states, obtaining the tuples in the lifted space, i.e., $(\bm{z}_0^j,\bm{u}_0^j)\in\mathcal{Z}^f_0\times\mathcal{U}$ and $({\bm{z}}_g^j,{\bm{u}}_{g}^j)\in{{{\mathcal{Z}}}}^b_0\times\mathcal{U}$.
Next, we define a time step size $\delta$, and forecast the initial observables with the consistent $\bm{u}_0^j$ by \eqref{eq:FINN} and backcast the goal observables with the ${\bm{u}}_{g}^j$ through \eqref{eq:BINN} over $T$ time period simultaneously.
We recover the predicted states from the generated observables. 
The convex hulls of the resulting states forward and backward propagation are denoted as $\text{Co}(\mathcal{X}^f_{0:T})$ and $\text{Co}({\mathcal{X}}^b_{-T:0})$, respectively.

In most cases, the convex polytopic sets $\mathcal{X}^f_0$ and ${\mathcal{X}}^b_0$ are mapped to nonconvex sets $\mathcal{Z}^f_0$ and ${{\mathcal{Z}}}^b_0$ since $\bm{z}(\bm{x})$ are nonlinear functions of $\bm{x}$.
Our convex hulls will be over-approximations, sufficient for a heuristic sampling.
To avoid returning a subset of the true and convex reachable set, we apply the adversarial sampling \cite{lew2021sampling} to inflate the resulting convex hulls.
To over-approximate FRS, after obtaining the set of the total $T$ period forward propagated trajectories $\mathcal{X}^f_{0:T}=\cup_{0\leq t\leq T}\{\bm{x}_{t}^j\}_{j=1}^M$, we use projected gradient ascent to maximize $\mathcal{L}^M(\bm{p})=\frac{1}{T}\sum_{t=\delta}^{T}\norm{\bm{x}_t(\bm{p}_0)-\bm{c}_t^M}^2_{Q_t^M}$ to regenerate initial tuples $\bm{p}_0=(\bm{x}_0^j,\bm{u}_0^j)$, where $Q_t^M$ are the positive definite matrices and $\bm{c}_t^M$ represent the geometric center of $\mathcal{X}^f_t$.
$\bm{p}_0\leftarrow \bm{p}_0+\eta\nabla_p\mathcal{L}^M(\bm{x}_{\delta:T}^j)$, where $\eta$ is a constant.
After projecting the new $p_0$ onto $\mathcal{X}^f_0\times\mathcal{U}$, we propagate the new samples again using our DIKU and recover the trajectories to obtain $\{\bm{x}_{0:T}^j\}_{j=1}^M$ that inflates $\text{Co}(\mathcal{X}^f_{0:T})$.
The propagated trajectories are updated as $\mathcal{X}^f_{0:T}\leftarrow\mathcal{X}^f_{0:T}\cup\{\bm{x}_{0:T}^j\}_{j=1}^M$.
% It is possible to conduct adversarial sampling several times if necessary, but one time usually works fine in practice.
According to Lemma \ref{lemma1}, the over-approximation of FRS at time $t$ is
$\mathcal{F}(0,t,\bm{x}_0)=\text{Co}(\mathcal{X}^f_t)$.

Similarly, we resample the goal tuples $p_g=(\bm{x}_g^j,\bm{u}_g^j)$ and backcast the $p_g$ by utilizing our backward DIKU and retrieve $\{{\bm{x}
}_{-T:0}^j\}_{j=1}^M$.
The adversarial sampling procedure of backward trajectories is  ${\mathcal{X}}^b_{-T:0}\leftarrow{\mathcal{X}}^b_{-T:0}\cup\{{\bm{x}
}_{-T:0}^j\}_{j=1}^M$.
Given a terminal time instant $t_f$, the BRT over the time interval $[t,t_f]$ is over-approximated as $\mathcal{R}_b(t,t_f,\mathcal{X}^g)=\text{Co}({\mathcal{X}}^b_{t-t_f:0})$.

\textcolor{black}{TIS is finally over-approximated with adequate number of samples $M$ as below according to \eqref{eq:tis}}
\begin{equation}
\Omega(t_f)=\!\!\!\mathop\bigcup\limits_{0\leq t\leq t_f}\!\!\!\text{Co}(\mathcal{X}^f_{t})
\cap\text{Co}({\mathcal{X}}^b_{t-t_f:0}).
\label{eq:convex_tis}
\end{equation}
Although we can encode the states in the sets $\mathcal{X}^f_0$ and ${\mathcal{X}}^b_{0}$, it is intractable to embed the sets that have arbitrary shape representations into $\mathcal{Z}^f_0$ and ${{\mathcal{Z}}}^b_{0}$ \cite{bak2021reachability}.
Besides, inflation errors would be added in the process due to the extra $\bm{z}_{\theta_z}(\bm{x})$ even though we predefine the shapes of $\mathcal{Z}^f_0$ and ${{\mathcal{Z}}}^b_{t_f}$ based on experience.
Thus, we only propagate the state but do not sample adversarially in the lifted space.
\textcolor{black}{System dynamics can be used directly with the sampling-based technique \cite{lew2021sampling} to obtain a better accuracy while the forward DIKU, along with the technique, guarantees online FRS approximation, even possible for some unknown systems.}

\section{Online Time-Informed \textcolor{black}{SKMP} of Nonlinear Systems}  
\begin{algorithm}[htbp]
	\normalem
	\caption{Online Time-Informed Sampling-based Kinodynamic Motion Planning}
	\KwIn{$(\bm{x}_0,\mathcal{X}^g);\;Q=\{\bm{x}_0\};\;E=\emptyset$}
	\KwOut{$Tree(Q,E)$}
	$\tau=0;\;\mathcal{X}^b_{0}=\mathcal{X}^g;\;M=1000;\;\eta;\;Tree(Q,E);$\\
	\While{$\bm{x}_0\notin\rm{Co}(\mathcal{X}^\textit{b}_{\tau})$}{
		$\tau=\tau-\delta$;
		% $\text{Co}(\mathcal{X}^b_{\tau:0})=\text{Co}(\mathcal{X}^b_{\tau+\delta:0})\cup\text{Co}(\mathcal{X}^b_{\tau})$;
	}
	$cost=-\tau$;\\
	% $\text{Co}(\mathcal{X}^b_{-cost:0})=\text{Co}(\mathcal{X}^b_{\tau+\delta:0})\cup\text{Co}(\mathcal{X}^b_{\tau})$;
	$\Omega(cost)=$ASKU$(cost,\bm{x}_0,\mathcal{X}^g,\eta,M)$;\Comment{Sec. \ref{sec:kotis}}\\
	$cost'=\infty$;\\
	% \text{Co}(\mathcal{X}^f_{0:cost}),\text{Co}(\mathcal{X}^b_{-cost:0})
	\While{$|(cost-cost')/cost'|>eps$}{
		$cost'=cost$;
		$[Tree,cost_{new}]=$TreeGrowth$(Tree,cost,\Omega(cost))$;
		\If{$cost_{new}\neq\infty$}{
			$cost=cost_{new}$;\\
			\If{$cost_{new}<cost$}{
				\begin{small}
					$\Omega(cost)=\Omega(cost')\backslash(\text{Co}(\mathcal{X}^f_{cost'})\cap\text{Co}({\mathcal{X}}^b_{0}))\backslash(\bigcup\limits_{0\leq t\leq cost+\delta}\text{Co}(\mathcal{X}^f_{t})\cap\text{Co}({\mathcal{X}}^b_{t-cost':t-cost-\delta}))$
				\end{small};
				$Tree=$Prune$(\Omega(cost),Tree,cost)$;
			}
		}
		\Else{
			$cost=cost+\delta_2$;
			\begin{small}
				$\Omega(cost)=\Omega(cost')\cup(\text{Co}(\mathcal{X}^f_{cost})\cap\text{Co}({\mathcal{X}}^b_{0}))\cup(\bigcup\limits_{0\leq t\leq cost-\delta}\text{Co}(\mathcal{X}^f_{t})\cap\text{Co}({\mathcal{X}}^b_{t-cost:t-cost'-\delta}))$
			\end{small};
		}
	}
	\label{alg:planning}
\end{algorithm}
\begin{figure*}[htbp]
\centering
\includegraphics[width=0.9\textwidth]{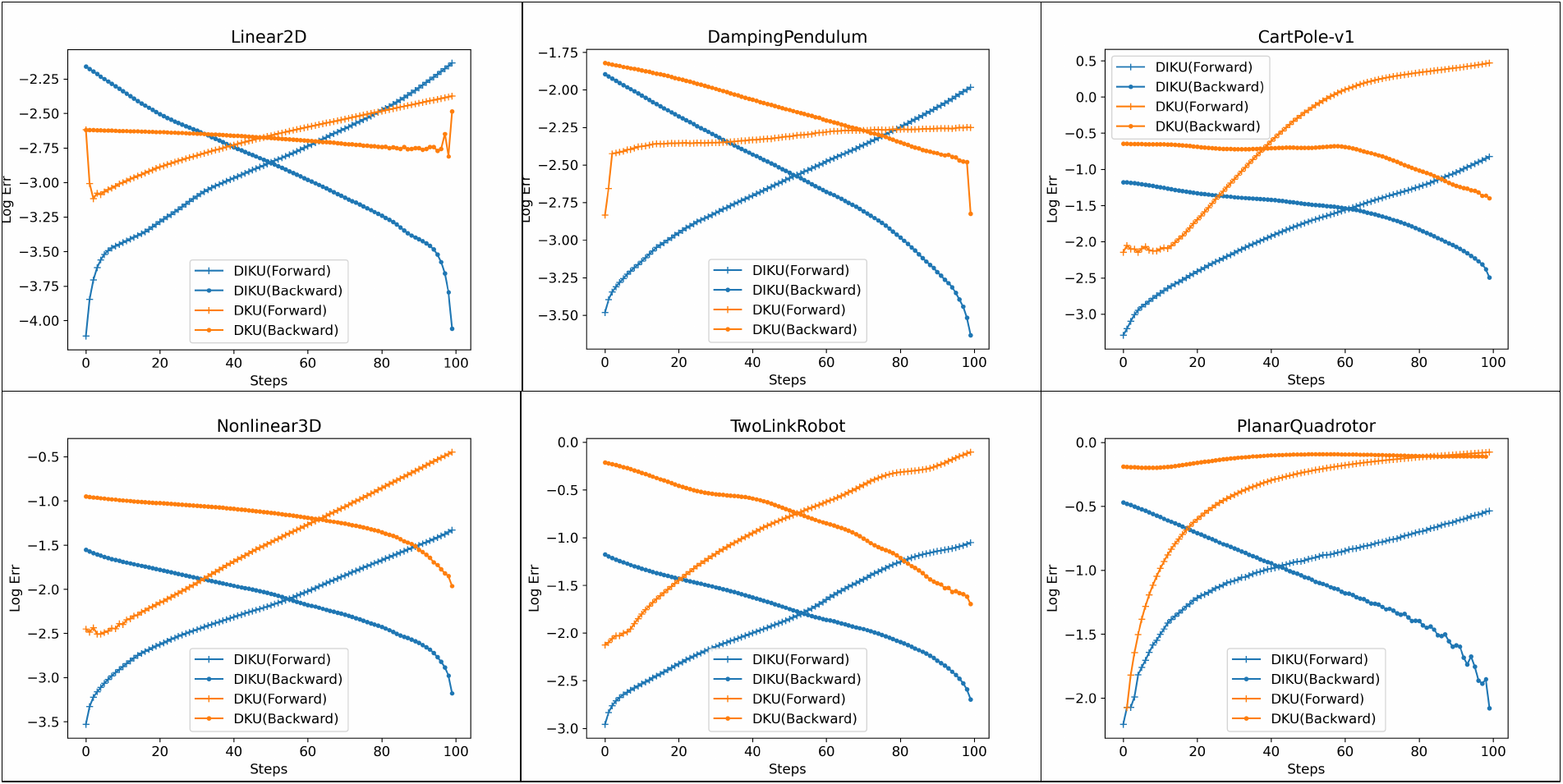}
\caption{\textcolor{black}{Comparison of forward and backward dynamics prediction by our DIKU and the DKU with consistency loss. The early predictions are more accurate than the late ones, which can be observed from the lines with plus markers of forward prediction from left to right and from those lines with dot markers of backward from right to left. The lower blue lines indicate smaller bidirectional dynamics evolution errors by DIKU, especially for nonlinear systems.}
}
\label{fig:prediction}
\end{figure*}
Algorithm $1$ outlines the procedure of our planning method.
We obtain the optimal time cost ignoring obstacles by determining the index of BRS that contains the start state $\bm{x}_0$.
The TIS corresponding to this $cost$ is constructed using our ASKU approach in Section \ref{sec:kotis}.
Then, we grow a search tree within $\Omega(cost)$ and compute the tree's time cost $cost_{new}$ while considering obstacles.
$\Omega(cost)$ reduces and we eliminate the nodes outside it once we find a better suboptimal solution.
In contrast, we extend the optimal time of the planning problem $\delta_2$s and expand the TIS since a collision-free solution less than $cost$ might not exist.

% It is challenging to directly draw uniform random samples in $\Omega(cost)$  due to the complexity of determining the reachable set intersection's shape and volume \cite{joshi2021tie}.
Algorithm $2$ describes the procedure to grow a search tree with a direct sampling technique, HNR sampling method \cite{10.1145/1921598.1921600}, within the components of TIS.
A time $t$ is first uniformly sampled in $[0,cost]$ as the cost-to-come of the sample to be generated.
If a random number is generated below the threshold $\mu$, we draw a sample within the TIS to accelerate the convergence speed; otherwise, within the entire state space.
% Unlike the time-consuming indirect sampling method in \cite{joshi2021tie,tang2022reachability}, we implement direct sampling .
We compute the longest time cost from $t$ to reach the goal.
A linear programming problem, subject to the linear inequalities of the FRS at $t$ and BRS at $t'$, is formulated to check if there exists any intersection.
% The objective function can be arbitrary since we're only interested in finding a feasible point.
We decrease the cost-to-go until finding the minimum value.
A valid cost-to-go $\Bar{t}$ is then uniformly sampled.
% The method involves starting from the Chebyshev center within the convex set and iteratively moving to new points by randomly selecting directions and lengths of movement within the set's constraints.
Finally, we use the vertex inclusion algorithm \cite{joshi2021tie} and an off-the-shelf SKMP to find a solution tree and return its cost $cost_{new}$, if there exists. 
% If fails after $N$ iterations, $cost_{new}=\infty$.
% \begin{equation}
% \omega(t,t_f)\!=\!\!\!\!\!\!\!\!\!\!\!\!
% \mathop\bigcup\limits_{\Bar{t}\in[t-t_f,\dots,-\delta,0]}\!\!\!\!\!\!\!\!\!\!\!\{\bm{x}\in\mathcal{X}\,|\,A^f_t\bm{x}+B^f_t\leq0, A^b_{\Bar{t}}\bm{x}+B^b_{\Bar{t}}\leq0
% \}.
% \end{equation}
% $\Omega(t_f)$ in \eqref{eq:tis} is further represented as
% \begin{equation}
% \Omega(t_f)=\!\!\!\!\!\!\mathop\bigcup\limits_{t\in[0,\delta,\dots,t_f]}\!\!\!\!\!\!\omega(t,t_f).
% \end{equation}

% disadvantage:The FRS corresponding to time t might not be found in the stored discrete reachable sets, e.g. t ∊ (-t k+1 , -t k ),
\section{Simulation Experiments}
In this section, we conduct simulation experiments to 1) verify our DIKU's bidirectional long-term prediction performance, 2) demonstrate the high performance of our ASKU's reachability analysis for TIS, and 3) show the improved planning efficiency of our online time-informed SKMP.
%\begin{algorithm}[t] 
%\normalem
%\KwIn{$Tree,cost,\Omega(cost)$}
%\KwOut{$Tree,cost_{new}$}
%\caption{Tree Growth} 
%\For{$i=1:N$}{
%$t=\text{UniformSample}([0,cost])$;\Comment{cost-to-come}
%\If{$\rm{Rand}(0,1)<\mu$}{
%$t'=t-cost$;\Comment{the maximum cost-to-go}\\
%% $\{\bm{x}\!\in\!\mathcal{X}|A^f_t\bm{x}\!+\!B^f_t\!\leq\!0,A^b_{{t}'}\bm{x}\!+\!B^b_{{t}'}\!\leq\!0\}\!\neq\!\emptyset$
%\While{$\{\bm{x}\!\in\!\mathcal{X}|\rm{Co}(\mathcal{X}^\textit{f}_\textit{t})\!\cap\!\rm{Co}(\mathcal{X}^\textit{b}_{\textit{t}'}),\textit{t}'\!\leq\!0\}\!\neq\!\emptyset$}{
%$t'=t'+\delta$;\Comment{find the minimum cost-to-go} \\
%}
%$\Bar{t}=\text{UniformSample}([t-cost,t'))$\;\\
%$\bm{x}_{new}$=HNR$(\{\bm{x}\in\mathcal{X}\,|\,\text{Co}(\mathcal{X}^f_t)\cap\text{Co}(\mathcal{X}^b_{\Bar{t}})\})$;
%}
%\Else{
%$\bm{x}_{new}=$Sample$(\mathcal{X})$;
%}
%$[Tree,cost_{new}]=\text{SKMP}(Tree,\bm{x}_{new},t)$;}
%\If{$\bf{no}$ node $\in\mathcal{X}^g$}{
%$cost_{new}=\infty$;}
%\label{alg:treegrow}
%\end{algorithm}  

\begin{algorithm}[t]
	\normalem
	\caption{Tree Growth} 
	\KwIn{$Tree,cost,\Omega(cost)$}
	\KwOut{$Tree,cost_{new}$}
	\For{$i=1:N$}{
		$t=\text{UniformSample}([0,cost])$; \Comment{cost-to-come}\\
		\If{$\rm{Rand}(0,1)<\mu$}{
			$t'=t-cost$; \Comment{the maximum cost-to-go}
			\While{$\{\bm{x}\!\in\!\mathcal{X}|\rm{Co}(\mathcal{X}^\textit{f}_\textit{t})\!\cap\!\rm{Co}(\mathcal{X}^\textit{b}_{\textit{t}'}),\textit{t}'\!\leq\!0\}\!\neq\!\emptyset$}{
				$t'=t'+\delta$; \Comment{find the minimum cost-to-go}
			}
			$\Bar{t}=\text{UniformSample}([t-cost,t'))$;
			$\bm{x}_{new}=$HNR$(\{\bm{x}\in\mathcal{X}\,|\,\text{Co}(\mathcal{X}^f_t)\cap\text{Co}(\mathcal{X}^b_{\Bar{t}})\})$;
		}
		\Else{
			$\bm{x}_{new}=$Sample$(\mathcal{X})$;
		}
		$[Tree,cost_{new}]=\text{SKMP}(Tree,\bm{x}_{new},t)$;
	}
	\If{$\bf{no}$ node $\in\mathcal{X}^g$}{
		$cost_{new}=\infty$;
	}
	\label{alg:treegrow}
\end{algorithm}

\subsection{Experiment Environments}
We have six dynamic systems for evaluation, including a (a) 2D linear system in \cite{tang2022reachability} (2D-L), (b) 3D polynomial nonlinear system in \cite{tang2022reachability} (3D-PNL), (c) CartPole in \cite{shi2022deep}, (d) DampingPendulum in \cite{shi2022deep}, (e) two-link robot in \cite{shi2022deep}, and (f) planar quadrotor in \cite{folkestad2021koopman}. \textcolor{black}{For each system, a total of 10000, 2500, and 5000 trajectories are generated by simulating the dynamics, designated for training, validation, and testing, respectively.}
Our experiments were conducted in Python and C++ using a 64-bit workstation with an Intel Core i9-13900K processor, NVIDIA 4090 GPU, and 128GB RAM, running Ubuntu 20.04 OS.

\subsection{Long-Term Forward and Backward Dynamics Prediction}
We evaluate the long horizon forward and backward dynamical prediction performance of our DIKU model on the six dynamics and compare them with the existing method.
We extend the Deep KoopmanU with Control (DKU) in \cite{shi2022deep} to compute the backward counterpart of \eqref{eq:kstepz} through additionally training an independent linear NN of $K_x^{-1}$.
We impose the weights of parameterized $K_x$ and $K_x^{-1}$ in a consistency loss as in \cite{azencot2020forecasting}.
One INN module is constructed, where $\bm{z}_{t,2}$ is randomly selected.
The hidden layers of $A_1$ and $A_2$ are $[128,64,32]$ with the ReLU \cite{agarap2018deep} in the dynamics (a)-(e), while with linear activation functions in (f).
The structures and weights of embedding and control nets, training and testing datasets, learning rate $0.001$, batch size $1024$, Adam optimizer \cite{kingma2014adam}, and K step loss with $\gamma=0.9$ of the extended DKU and DIKU are identical for each system. 
% DIKU and the baseline share the same weights of embedding and control nets to highlight the prediction improvement brought by our novel DKO model design.
% We use the hyperparameters, encoder, and loss definition of \cite{folkestad2021koopman} in (f) to maintain a fair comparison.

The 100-step bidirectional prediction results are shown in Fig. \ref{fig:prediction}. 
We calculate the mean of the log10 of the maximum error.
It can be seen that our DIKU (blue) can bidirectionally predict much more accurate states in the long horizon than the DKU (yellow).
Unlike the DKU which uses different structures, parameters, and a soft loss function to train $K_x$ and $K_x^{-1}$ for bidirectional prediction, our model allows direct back prediction through the INN structure \eqref{eq:BINN},
benefiting a better inference and convergence in training.
% We have tried different hidden layers and linear activation functions as in \cite{tayal2023koopman} of INN but the results were not greatly improved.
Another reason is that the control field's nonlinearity can be reflected by the INN's translation functions.
% We randomly choose the elements of $\bm{z}$ to form $\bm{z}_{t,1}$ and generally design and train the INN in our DIKU. 
%Better prediction results could be gained if we take good care of choosing  $\bm{z}_{t,2}$.
Consequently, DIKU paves the way to efficiently perform accurate long-term bidirectional propagation in sampling-based reachability analysis.

\subsection{Online Time-Informed Set Approximation}
\label{sec:onlinetisapp}
\begin{figure}[htbp] 
  \centering 
  \subfigure[Forward Reachable Sets and Tubes]{ 
    %\label{fig:subfig:a} %% label for first subfigure 
    \includegraphics[width=0.45\textwidth,height=0.45\textwidth]{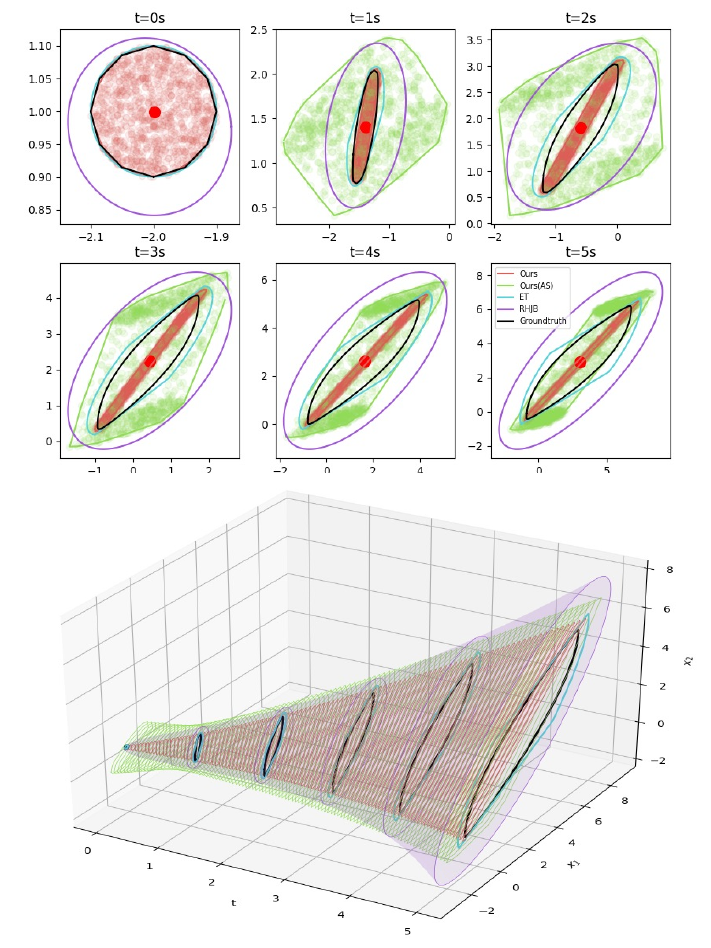} 
  } 
  \subfigure[Backward Reachable Sets and Tubes]{ 
    %\label{fig:subfig:b} %% label for second subfigure 
    \includegraphics[width=0.45\textwidth,height=0.45\textwidth]{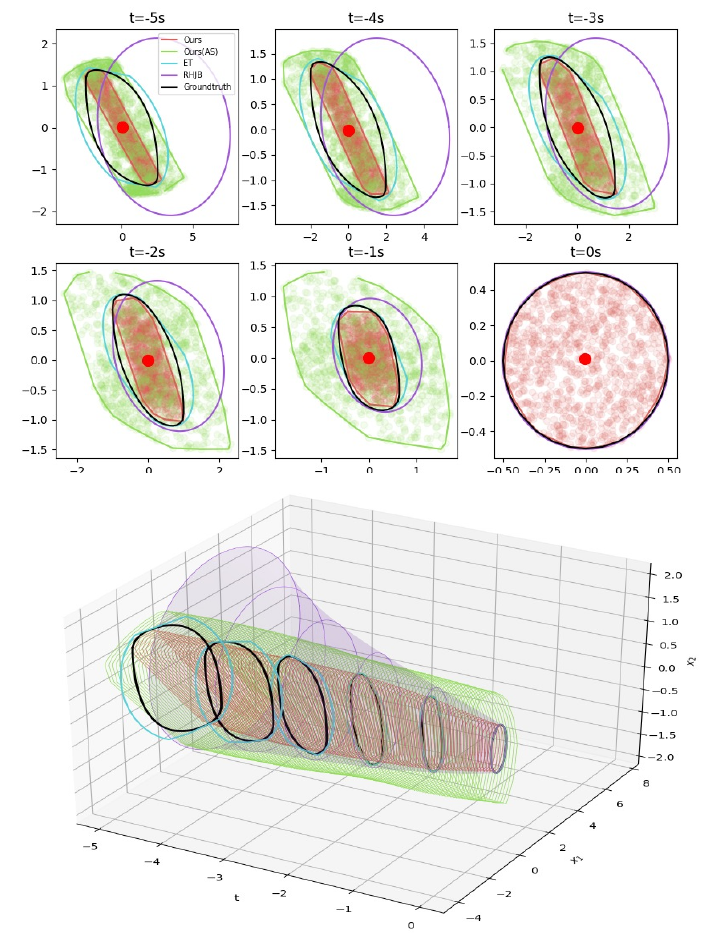} 
  } 
  \caption{Comparison results of the example 2D system \cite{tang2022reachability} (a) forward reachable sets and tubes and (b) the backward counterparts of TIS computed by the level set toolbox \cite{mitchell2005time} (Ground truth, black), ellipsoidal toolbox \cite{kurzhanski2000ellipsoidal} (ET, cyan), relaxed HJB equation method \cite{tang2022reachability} (RHJB, purple), our basic sampling-based convex approximation (Ours, red), and our adversarial sampling for over-approximation, ASKU, (Ours (AS), green). Our method provides inner-approximated and tight over-approximated TISs online.} 
  \label{fig:tis}
\end{figure}

\begin{figure}[htbp]
\centering
\includegraphics[width=0.45\textwidth]{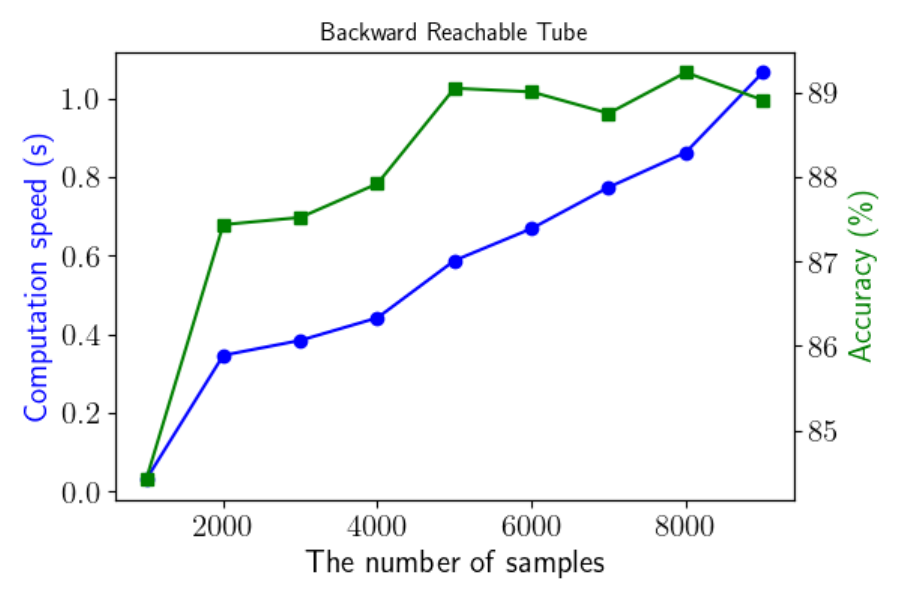}
\caption{\textcolor{black}{Volume convergence (green line) and speed (blue line) on DIKU with a varying number of samples $M$ for our basic BRT approximation. The volume accuracy is computed by being divided by the volume of ground truth. A high degree of accuracy can be obtained at the cost of computation speed.}
}
\label{fig:samples}
\end{figure}

We compare our basic sampling-based approach (ours) and its over-approximation version, the ASKU, (ours(AS)) with level set method \cite{mitchell2005time} (ground truth), ellipsoidal toolbox \cite{kurzhanski2000ellipsoidal} (ET), and relaxed HJB equation method \cite{tang2022reachability} (RHJB) for the dynamics (a). 
Set identical time step $0.05$s and horizon $T=5$s. 
\textcolor{black}{We evaluate the volume convergence of ours for computing BRT and show the results in Fig. \ref{fig:samples}.}
We set $M=1000$, $\eta=0.015$, and $\eta=0.04$ for forward and backward prediction
% $M$ is determined to balance the volume of convex TIS and computation speed and 
to let the gradient and $\bm{p}_0$ have the same order of magnitude.
We use the ellipsoidal parameters of initial and goal sets and control limits in \cite{tang2022reachability}. 
The reachable sets and tubes that constitute TIS are shown in Fig. \ref{fig:tis}.
Since the ground truth has to discretize the spaces to calculate the numerical result of the HJB partial differential equation, we have tried to reduce the number of girds to save time while guaranteeing the quality of the reachable sets.
Table \ref{tab:TIStime} records the computation costs of reachable tubes.
The ground truth takes plenty of time to return a fine FRT.
ET must be calculated in 30 random directions, resulting in a heavy computation burden and a more compact ellipsoid than RHJB.
RHJB predefines the centers of the external ellipsoids 
instead of treating them as decision variables since that makes solving SOS optimization intractable.
However, the centers highly influence the convergence of RHJB, leading to no solution sometimes.
Although sometimes feasible for the system, it is time-consuming and may cause drift, as seen in the purple BRT in Fig. \ref{fig:tis}.
These methods also suffer from the curse of dimensionality.
In contrast, ours and ASKU can be completed in near real-time. 
Note that ASKU can be applied to linear and nonlinear systems that can be learned regardless of the number of dimensions, thanks to DIKU's NN bidirectional propagation.
% Note that only two parameters $\eta$ and $M$ exist in our method to realize online TIS approximation.
Facing the true convex reachable set, our ASKU can still return over-approximation online.
%The reachable sets might be conservative at some time steps since we consider the entire period in \eqref{eq:gradient loss}.
%However, conservative TIS is conducive to guaranteeing the probabilistic completeness of informed SKMP to some extent. 
\subsection{Online Time-Informed Kinodynamic Motion Planning}
\begin{table}[htbp]
\centering
\caption{Computation times \textcolor{black}{[s]} of the Forward/Backward reachable tubes for the example 2D system in \cite{tang2022reachability}}
\label{tab:TIStime}
\begin{tabular}{cccccc}
    \toprule
    & Level Set \cite{mitchell2005time} & ET \cite{kurzhanski2000ellipsoidal} & RHJB \cite{tang2022reachability} & Ours & Ours (AS)\\
    \hline
FRT & 47.46 & 51.13 &22.17& \textbf{0.03}&0.16\\
BRT & 5.71  & 6.07  &23.31& \textbf{0.03}&0.16\\
    \bottomrule
\end{tabular}
\end{table}
We conduct the \textcolor{black}{problem \eqref{eqoptimalcost}} on all dynamics between SST \cite{li2016asymptotically} and our algorithm while only comparing the approach in 
% \begin{figure}[h]
% \centering
% \includegraphics[width=0.48\textwidth]{./figures/cost.png}
% \caption{The statistical comparison of the costs (red) of planned solutions and the corresponding sampled numbers of nodes (blue) in six environments.
% }
% \label{fig:cost}
% \end{figure}
\cite{tang2022reachability} with default parameters on (a) and (b). 
We set $\delta_2=0.5$s, and $M=1000$ and valid $\eta$ for our method.
Set $\mu=0.9$ to accelerate our search progress and $N=500$ in \cite{tang2022reachability} and ours. \textcolor{black}{A library of TIS computed offline exists only for \cite{tang2022reachability}. The running time of approximating TIS is not counted in \cite{tang2022reachability} but it is included in ours because ours is an online time-informed method. The computation time is measured as the condition (line 7, Alg. 1) is no longer satisfied.}

\begin{figure*}[thbp!]
    \centering
    \subfigure[SST (left): $T_{OP}=\SI{23.18}{s}$, $C_{OP}=\SI{4.93}{s}$; Ours (right): $T_{OP}=\SI{6.76}{s}$, $C_{OP}=\SI{4.49}{s}$.]{ 
    \includegraphics[width=0.45\textwidth,height=0.15\textwidth]{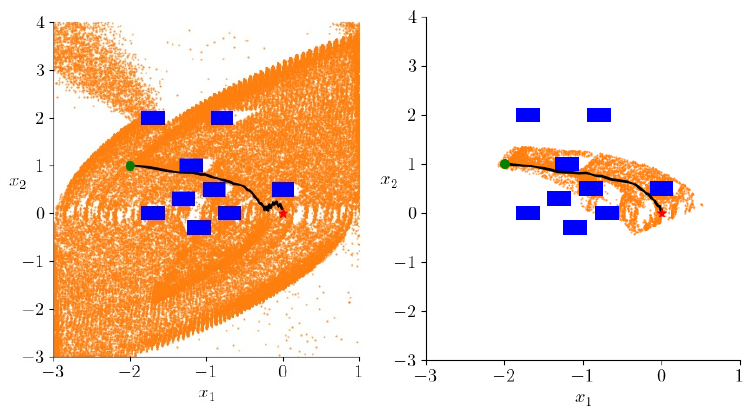} 
    } 
    \subfigure[SST (left): $T_{OP}=\SI{55.87}{s}$, $C_{OP}=\SI{5.56}{s}$; Ours (right): $T_{OP}=\SI{4.49}{s}$, $C_{OP}=4.04$.]{ 
    \includegraphics[width=0.45\textwidth,height=0.15\textwidth]{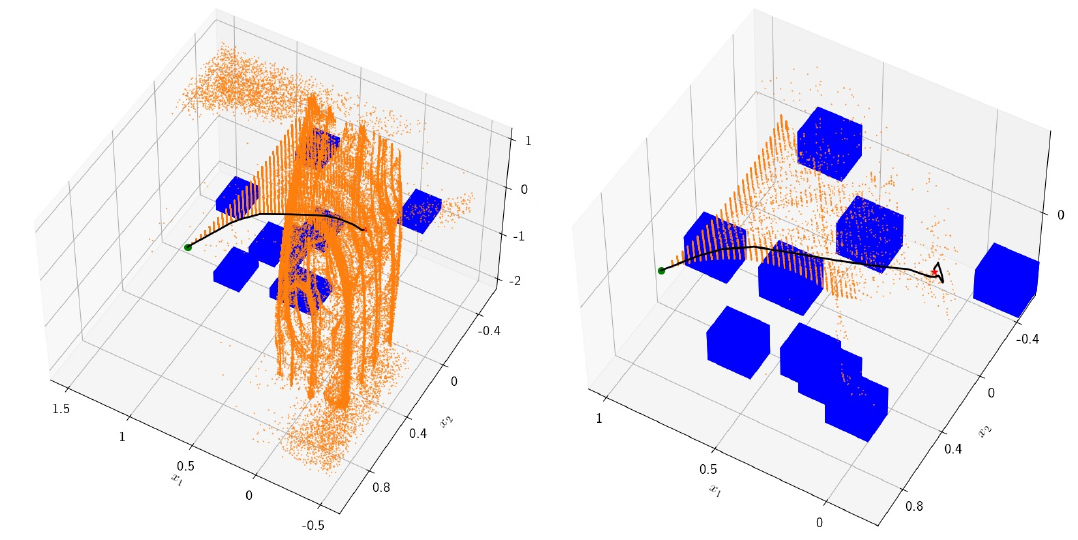} 
    }
    \subfigure[SST (left): $T_{OP}=\SI{28.79}{s}$, $C_{OP}=\SI{6.27}{s}$; Ours (right): $T_{OP}=\SI{12.21}{s}$, $C_{OP}=\SI{5.57}{s}$.]{ 
    \includegraphics[width=0.45\textwidth,height=0.15\textwidth]{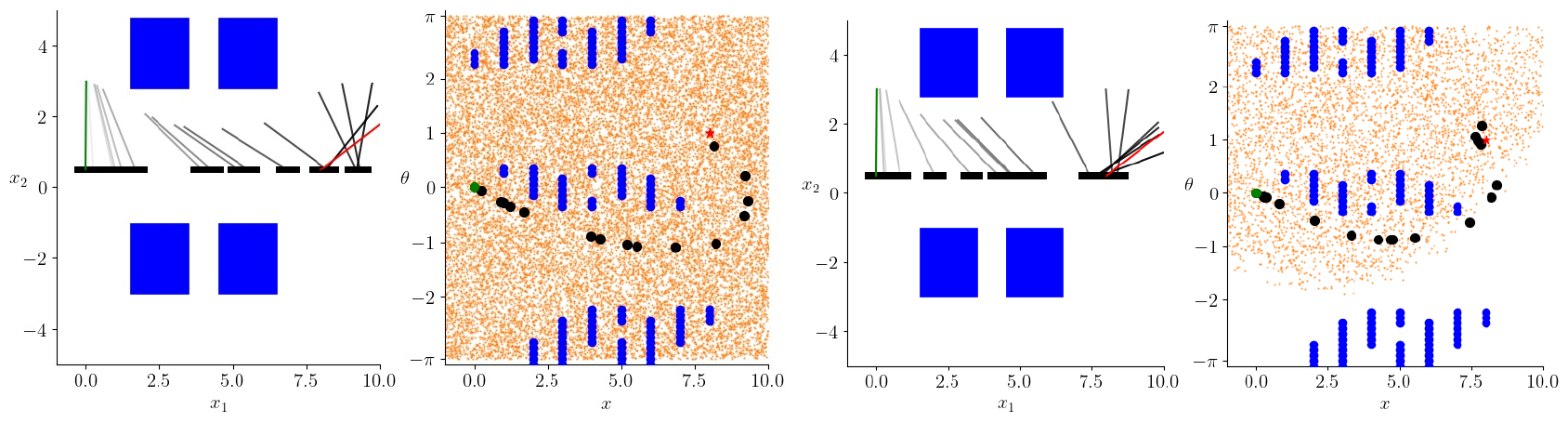} 
    } 
    \subfigure[SST (left): $T_{OP}=\SI{33.15}{s}$, $C_{OP}=\SI{1.89}{s}$; Ours (right): $T_{OP}=\SI{10.48}{s}$, $C_{OP}=\SI{1.86}{s}$.]{ 
    \includegraphics[width=0.45\textwidth,height=0.15\textwidth]{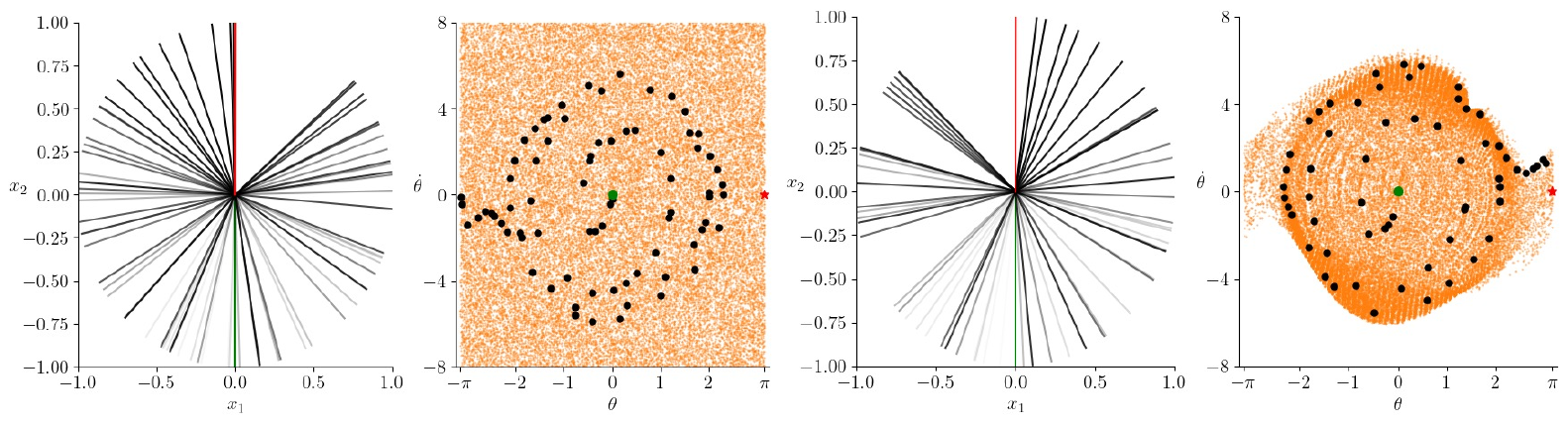} 
    }
    \subfigure[SST (left): $T_{OP}=\SI{46.22}{s}$, $C_{OP}=\SI{4.54}{s}$; Ours (right): $T_{OP}=\SI{18.15}{s}$, $C_{OP}=\SI{4.35}{s}$.]{ 
    \includegraphics[width=0.45\textwidth,height=0.15\textwidth]{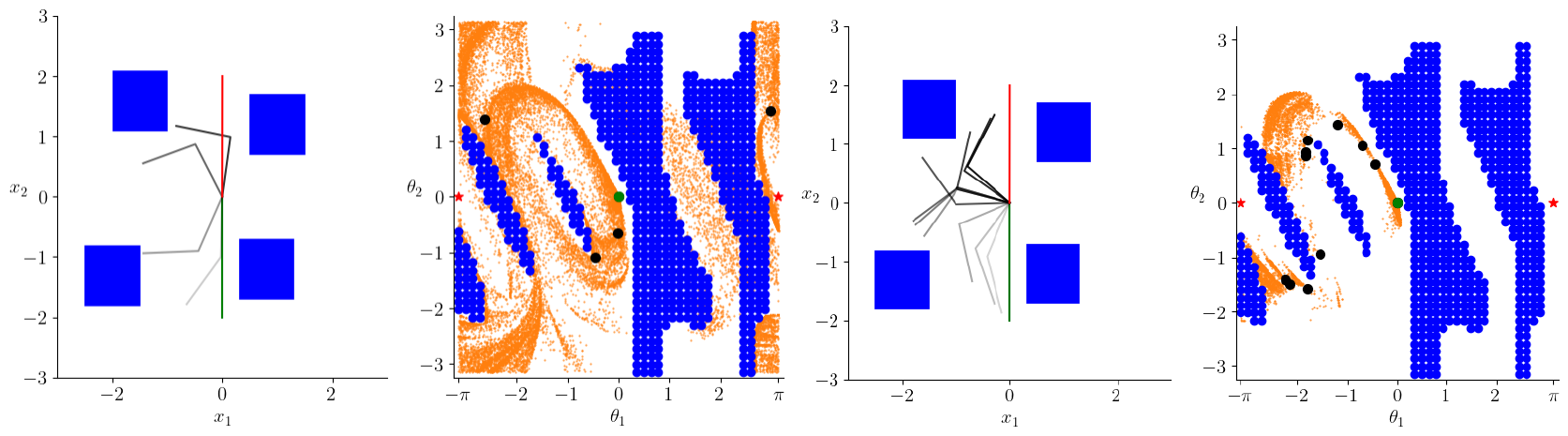} 
    } 
    \subfigure[SST (left): $T_{OP}=\SI{56.81}{s}$, $C_{OP}=\SI{9.16}{s}$; Ours (right): $T_{OP}=\SI{23.17}{s}$, $C_{OP}=\SI{8.86}{s}$.]{ 
    \includegraphics[width=0.45\textwidth,height=0.15\textwidth]{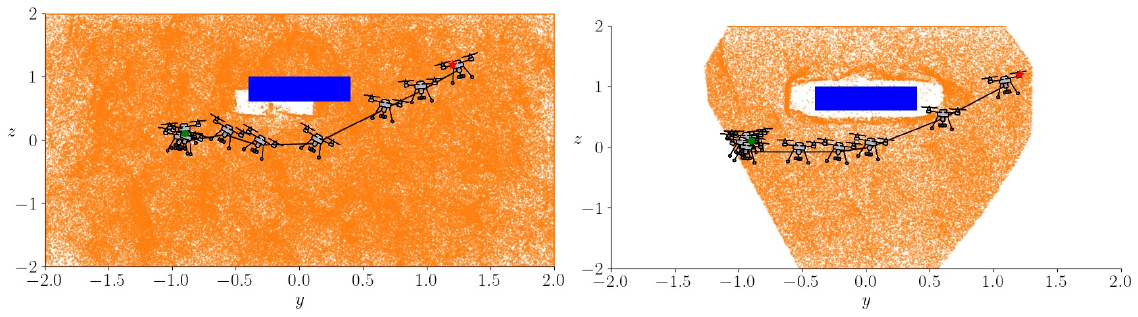} 
    }
\caption{Planning solutions (black lines) of the (a) 2D-L, (b) 3D-PNL, (c) CartPole, (d) DampingPendulum, (e) Two-link acrobot, and (f) planar quadrotor systems facing obstacles (blue boxes) given a start (green point) and goal (red pentagram) states. For (c)-(e), both the workspace and state-space trajectories are shown for each method. The orange points are samples. It can be seen that our online time-informed SKMP has a restricted search domain, indicating our ASKU is generalizable and scalable to nonlinear systems and benefits heuristic sampling. The planning efficiency is then improved.}
    \label{fig:problem}
 \end{figure*}
 
 \begin{table}
\renewcommand\arraystretch{1.2}
\scriptsize
\centering
\caption{Comparison results of the running time, path qualities, and the number of nodes for benchmarks and ours} 
\label{tab2}
\setlength\tabcolsep{1.6pt}
\begin{tabular}{ccccccc}
\toprule
&&$T_{IN}[s]$ & $C_{IN}[s]$ & $T_{OP}[s]$ & $C_{OP}[s]$ & $N_{node}(\times10^2)$\\ 
\hline
\multirow{3}*{2D-L} & Ours &$\bm{0.87}$&$\bm{4.68\!\!\pm\!\!0.10}$&$\bm{6.87}$&$\bm{4.44\!\!\pm\!\!0.09}$ &$\bm{3.21\!\!\pm\!\!0.7}$\\
& \cite{tang2022reachability} &$1.02$&$4.78\!\!\pm\!\!0.12$&$7.02$&$4.46\!\!\pm\!\!0.12$&$3.4\!\!\pm\!\!0.81$\\
& SST &$1.78$&$9.11\!\!\pm\!\!2.91$&$25.78$&$4.87\!\!\pm\!\!0.60$ &$12.9\!\!\pm\!\!0.29$\\
\hline
\multirow{3}*{3D-PNL} & Ours &$\bm{1.45}$&$\bm{4.22\!\!\pm\!\!0.21}$&$\bm{4.45}$&$\bm{4.07\!\!\pm\!\!0.08}$ &$\bm{1.67\!\!\pm\!\!0.27}$\\
& \cite{tang2022reachability} &$1.68$&$4.31\!\!\pm\!\!0.22$&$4.68$&$4.08\!\!\pm\!\!0.10$ &$1.71\!\!\pm\!\!0.28$\\
& SST &$26.37$&$6.38\!\!\pm\!\!2.08$&$56.37$&$5.33\!\!\pm\!\!1.40$ &$28.51\!\!\pm\!\!7.58$\\
\hline
\multirow{2}*{CartPole} & Ours &$\bm{5.26}$&$\bm{4.91\!\!\pm\!\!1.74}$&$\bm{11.26}$&$\bm{4.18\!\!\pm\!\!1.73}$ &$\bm{5.87\!\!\pm\!\!1.91}$\\
& SST &$14.69$&$8.21\!\!\pm\!\!4.90$&$29.69$&$6.15\!\!\pm\!\!4.21$ &$19.44\!\!\pm\!\!5.59$\\
\hline
Damping & Ours &$\bm{6.23}$&$\bm{2.08\!\!\pm\!\!0.35}$&$\bm{10.57}$&$\bm{1.85\!\!\pm\!\!0.39}$ &$\bm{5.26\!\!\pm\!\!1.78}$\\
Pendulum & SST &$21.2$&$2.38\!\!\pm\!\!1.31$&$33.2$&$1.91\!\!\pm\!\!0.96$ &$21.25\!\!\pm\!\!6.51$\\
\hline
Two-link & Ours &$\bm{8.34}$&$\bm{4.72\!\!\pm\!\!1.17}$&$\bm{17.34}$&$\bm{4.39\!\!\pm\!\!1.56}$ &$\bm{6.52\!\!\pm\!\!2.03}$\\
Robot & SST &$27.12$&$7.10\!\!\pm\!\!5.20$&$45.12$&$4.60\!\!\pm\!\!4.60$ &$26.35\!\!\pm\!\!12.75$\\
\hline
Planar & Ours &$\bm{15.64}$&$\bm{8.50\!\!\pm\!\!1.02}$&$\bm{24.64}$&$\bm{7.72\!\!\pm\!\!1.12}$ &$\bm{45.54\!\!\pm\!\!13.04}$\\
Quadrotor & SST &$39.91$&$10.52\!\!\pm\!\!2.83$&$57.91$&$8.90\!\!\pm\!\!2.62$ &$234.03\!\!\pm\!\!52.21$\\
\bottomrule
\end{tabular}
\end{table}

Fig. \ref{fig:problem} depicts the schematics of the planning problems and Table \ref{tab2} reports the statistical results of 100 trials.
$T_{IN}$ and $T_{OP}$ represent the computation time of discovering the initial solution and final suboptimal solution, respectively. $C_{IN}$ and $C_{OP}$ stand for the solutions' costs, respectively. $N_{node}$ denotes the total number of nodes at the conclusion.
Since our method attempts to grow the search tree within $\Omega$ from the beginning, better initial and final collision-free solutions are returned much faster than SST in all environments.
It can be observed that our algorithm finds better initial and final solutions more quickly compared with \cite{tang2022reachability} because we directly draw samples in a tighter TIS.
% Note that our method does not need to take a large amount of time to compute the TIS before it commences and does not worry that there might be an absence in the stored discrete reachable sets during online planning.
It can also apply to the nonlinear systems (c)-(f) other than the PNL system (b).
\textcolor{black}{The sampling acceptance rate among the six systems is $82\%$ on average.}
In terms of memory burdens, our algorithm and \cite{tang2022reachability} require fewer nodes than SST in the environments of (a) and (b) due to the elimination effect of the Prune() function.
In the other environments, our method retains the tree with $19.5\%\sim30.2\%$ nodes since we sample within the $\Omega(cost)$ that contains the optimal solution.

\section{CONCLUSION}
In this work, we realize online TIS approximation and propose an online time-informed SKMP of nonlinear systems.
Our proposed Deep Invertible Koopman operator with control U (DIKU) model accurately infers forward and backward dynamics propagation over a long horizon by designing the INN-based auxiliary network.
The developed ASKU method can over-approximate TIS in convex sets for a variety of systems at little cost through conducting adversarial sampling for DIKU bidirectional propagation.
The time-optimal SKMP that directly samples in the TIS takes less planning time than the \cite{tang2022reachability} algorithm although counting the TIS computation costs.
Our next research direction is to improve our DIKU and realize bidirectional search tree growth.

% \addtolength{\textheight}{-12cm}   % This command serves to balance the column lengths
                                  % on the last page of the document manually. It shortens
                                  % the textheight of the last page by a suitable amount.
                                  % This command does not take effect until the next page
                                  % so it should come on the page before the last. Make
                                  % sure that you do not shorten the textheight too much.

%%%%%%%%%%%%%%%%%%%%%%%%%%%%%%%%%%%%%%%%%%%%%%%%%%%%%%%%%%%%%%%%%%%%%%%%%%%%%%%%

%%%%%%%%%%%%%%%%%%%%%%%%%%%%%%%%%%%%%%%%%%%%%%%%%%%%%%%%%%%%%%%%%%%%%%%%%%%%%%%%

%%%%%%%%%%%%%%%%%%%%%%%%%%%%%%%%%%%%%%%%%%%%%%%%%%%%%%%%%%%%%%%%%%%%%%%%%%%%%%%%
%\section*{APPENDIX}

%Appendixes should appear before the acknowledgment.

%\section*{ACKNOWLEDGMENT}

%The preferred spelling of the word ÒacknowledgmentÓ in America is without an ÒeÓ after the ÒgÓ. Avoid the stilted expression, ÒOne of us (R. B. G.) thanks . . .Ó  Instead, try ÒR. B. G. thanksÓ. Put sponsor acknowledgments in the unnumbered footnote on the first page.

%%%%%%%%%%%%%%%%%%%%%%%%%%%%%%%%%%%%%%%%%%%%%%%%%%%%%%%%%%%%%%%%%%%%%%%%%%%%%%%%

\normalem
\bibliographystyle{unsrt}
\bibliography{reference}

\end{document}